\documentclass[submission]{eptcs}
\providecommand{\event}{CREST 2017}
\usepackage{breakurl}             

\usepackage{amstext}
\usepackage{amssymb}
\usepackage{todonotes}
\usepackage{url}
\usepackage{enumerate}
\usepackage{algorithmic}
\usepackage{multirow}
\usepackage{algorithm}
\usepackage{graphicx}
\usepackage{xspace}
\usepackage{subfig}
\usepackage[figuresright]{rotating}

\usepackage{tikz}
\usetikzlibrary{positioning, automata, arrows, calc, shapes, patterns}

\newcommand\ignore[1]{{}}

\newcommand{\ornode}{\texttt{OR}\xspace}
\newcommand{\andnode}{\texttt{AND}\xspace}

\newcommand\dLocal{Trace}
\newcommand\dGlobal{System}

\newcommand{\completeness}[1]{\colorbox{yellow!40}{$\displaystyle #1$}}
\newcommand{\correctness}[1]{\colorbox{orange!40}{$\displaystyle #1$}}
\newcommand{\diagnosability}[1]{\colorbox{cyan!40}{$\displaystyle #1$}}
\newcommand{\maximality}[1]{\colorbox{red!40}{$\displaystyle #1$}}

\newcommand{\strong}[1]{\ensuremath{{}_\llcorner\hspace*{-0.5mm}{#1}\hspace*{-0.5mm}{}_\lrcorner}}
\newcommand{\trigger}[1]{\ensuremath{\strong{#1}}}

\title{%
  Causality and Temporal Dependencies in the Design of Fault
  Management Systems }

\author{Marco Bozzano
\institute{Fondazione Bruno Kessler}
\email{bozzano@fbk.eu}
}

\def\titlerunning{Causality and Temporal Dependencies in the Design of Fault
  Management Systems}
\def\authorrunning{M. Bozzano}

\begin{document}
\maketitle

\begin{abstract}
  Reasoning about causes and effects naturally arises in the
  engineering of safety-critical systems. A classical example is Fault
  Tree Analysis, a deductive technique used for system safety
  assessment, whereby an undesired state is reduced to the set of its
  immediate causes. The design of fault management systems also
  requires reasoning on causality relationships. In particular, a
  fail-operational system needs to ensure timely detection and
  identification of faults, i.e. recognize the occurrence of run-time
  faults through their observable effects on the system.  Even more
  complex scenarios arise when multiple faults are involved and may
  interact in subtle ways.

  In this work, we propose a formal approach to fault management for
  complex systems. We first introduce the notions of fault tree and
  minimal cut sets. We then present a formal framework for the
  specification and analysis of diagnosability, and for the design of
  fault detection and identification (FDI) components.  Finally, we
  review recent advances in fault propagation analysis, based on the
  Timed Failure Propagation Graphs (TFPG) formalism.
\end{abstract}

\section{Introduction}

Modern complex engineering systems, such as satellites, airplanes and
traffic control systems need to be able to handle faults. Faults may
cause failures, i.e. conditions such that particular components or
larger parts of a system are no longer able to perform their required
function.  As a consequence, faults can compromise system safety,
creating a risk of damage to the system itself or to the surrounding
infrastructure, or even a risk of harm to humans.

For these reasons, complex system implement fault management systems.
There are different ways to deal with faults. {\em Fault avoidance}
tries to prevent design faults, through rigorous development
methodologies. Not all faults, however, can be prevented,
e.g. hardware faults may happen due to wear-out of components. {\em
  Fault tolerance}, on the other hand, aims at making a system robust
to faults that may occur during system operation, by using, e.g., a
redundant architecture, and by replicating critical components. A
fault tolerant system often implements some mechanisms to detect,
identify and recover from, faults -- i.e. an FDIR (Fault Detection,
Identification and Recovery) sub-system. In all cases, the design of
complex systems requires evaluating and quantifying the likelihood and
the consequences of failures. This process is called {\em safety
  assessment}. Classical techniques for safety assessment include
Fault Tree Analysis (FTA) and Failure Modes and Effects Analysis
(FMEA)~\cite{FTH2}.

In this paper, we introduce and review some recent work on
the design of fault management systems.  Our work leverages the use of
model-based design methodologies and formal verification and
validation techniques based on model
checking~\cite{clarke_mc,katoen_mc}.  We begin by introducing Fault
Tree Analysis in Sect.~\ref{sec:fta}. We then present a formal
framework for the design of fault detection and identification
components, in Sect.~\ref{sec:fdi}. We discuss techniques to analyze
fault propagation, based on the so-called {\em Timed Failure
  Propagation Graphs} (TFPGs) formalism, in Sect.\ref{sec:tfpg}.
Finally, we conclude in Sect.\ref{sec:conclusions} by outlining some
future directions.

\section{Fault Tree Analysis}
\label{sec:fta}
\begin{figure}[t]
\centering
\fbox{
\includegraphics[width=0.90\textwidth]{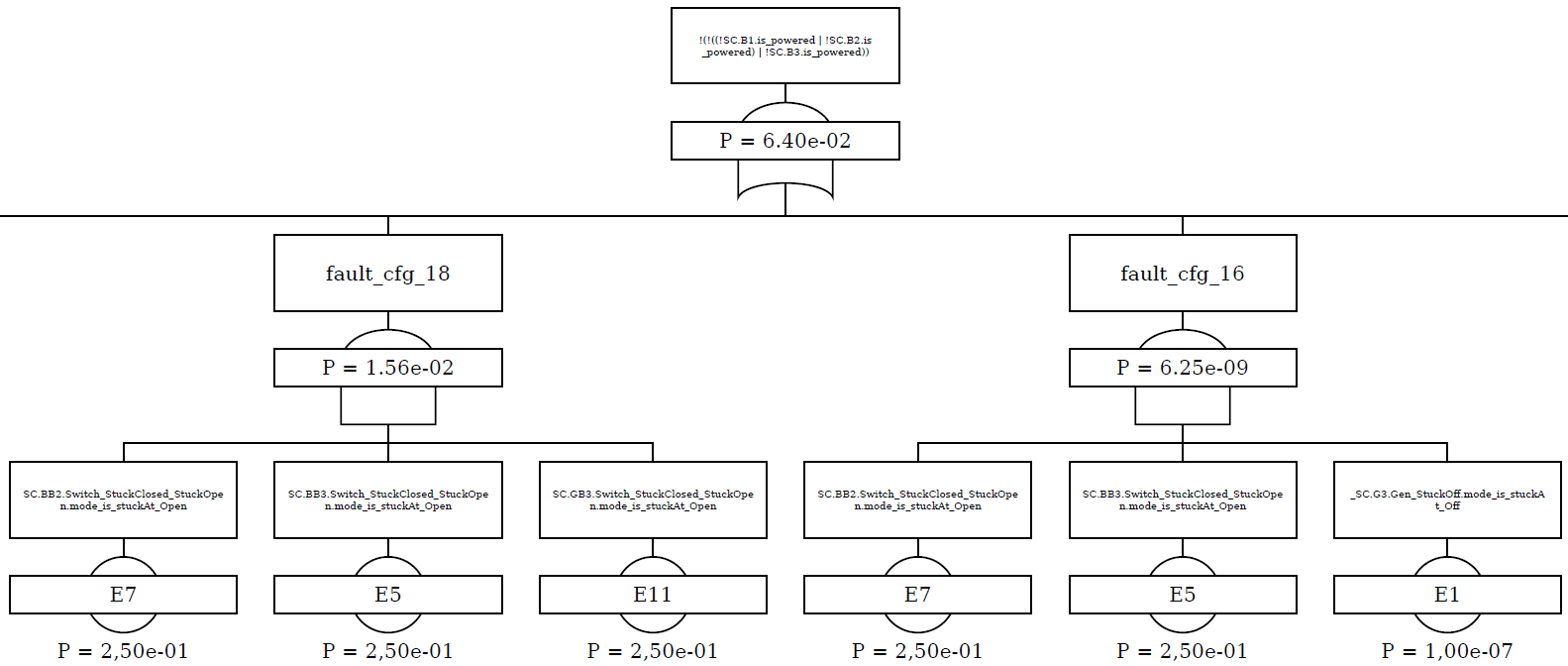}
}
\caption{An example fault tree.}
\label{fig:ft}
\end{figure}

Fault Tree Analysis (FTA)~\cite{FTH2} is a classical technique for
safety assessment. It is a deductive technique, whereby an undesired
state (the so called {\em top level event} (TLE) or {\em feared
  event}) is specified, and the system is analyzed for the possible
fault configurations (sets of faults, a.k.a. basic events) that may
cause the top event to occur. Fault configurations are arranged in a
tree, which makes use of logical gates to depict the logical
interrelationships linking such events with the TLE, and which can be
evaluated quantitatively, to determine the probability of the TLE.  An
example fault tree is shown in Figure~\ref{fig:ft}.

Of particular importance in safety analysis is the list of {\em
  minimal} fault configurations, i.e. the {\em Minimal Cut Sets
  (MCSs)}. More specifically, a cut set is a set of faults that
represents a necessary, but not sufficient, condition that may cause a
system to reach the top level event. Moreover, minimality implies that
every proper sub-set of a MCS is not itself a cut set.

Our approach to FTA is based on formal techniques, and specifically on
Model-Based Safety Analysis
(MBSA)~\cite{bozzano:03,ONERA2,bozzano:04,Joshi05:Dasc,ATVA07,SafetyAssessmentBook,DBLP:journals/cj/BozzanoCKNNR11}.
MBSA automates complex and error-prone activities such as the
generation of MCSs. This is done by looking for minimal fault
assignments, in symbolic models where selected variables represent the
occurrence of faults~\cite{ATVA07}. Cut sets are assignments to such
variables that lead to the violation of the top level event. Formal
verification tools for MBSA include xSAP~\cite{xSAP}.

Recent work~\cite{CAV-algos-paper} further improves existing
algorithms, by providing a fully automated generation of MCSs based on
state-of-the-art IC3 techniques~\cite{DBLP:conf/vmcai/Bradley11}. The
approach is {\em anytime}, in that it is able to compute an
approximation (lower bound and upper bound) of the set of MCSs, by
generating them for increasing cut set cardinality. This approach in
inspired by the layering approach
of~\cite{DBLP:conf/isola/AbdullaDSAA04}, but it improves over it in
several respects, such as scalability and convergence.  The approach
builds upon IC3-based parameter synthesis~\cite{FMCAD13}, by
providing several enhancements based on the specific features of the
problem.

\section{A Formal Framework for FDI}
\label{sec:fdi}
Fault Detection and Identification (FDI) is carried out by dedicated
modules, called FDI components, running in parallel with the system.
The detection task is the problem of understanding whether a
component has failed, whereas the identification task aims to
understand exactly which fault occurred. In general, detection and
identification may also apply to conditions other than faults.

Typically, faults are not directly observable and their occurrence can
only be inferred by observing the effects that they have on the
observable parts of the system. An FDI component processes sequences
of observations (made available by sensors) and triggers a set of
alarms in response to the occurrence of faults.

A formal foundation to support the design of FDI has been described
in~\cite{TACAS14,LMCS}, where a \emph{pattern based} language for the
specification of FDI requirements is proposed. \cite{TACAS14} focuses
on synchronous systems, whereas~\cite{LMCS} extends the framework to
include the asynchronous composition of the system with the diagnoser.

The first ingredient for specifying an FDI requirement is given by the
condition to be monitored, called {\em diagnosis condition}. The
second ingredient is the relation between the diagnosis condition and
the raising of an alarm. An {\em alarm condition} is composed of two
parts: the diagnosis condition and the delay. The delay relates the
time between the occurrence of the diagnosis condition and the raising
of the corresponding alarm. The language supports various forms of
delay: exact (\textsc{ExactDel}, after exactly $n$ steps), bounded
(\textsc{BoundDel}, within $n$ steps) and finite (\textsc{FiniteDel},
eventually).

The framework supports further aspects that are important for the
specification of FDI requirements. The first one is the
\emph{diagnosability} ~\cite{Sampath96}, i.e., whether the sensors
convey enough information to detect the required conditions.  A
non-diagnosable system (with respect to a given property) is such that
no diagnoser exists, that is able to diagnose the property.  The above
definition of diagnosability might be stronger than necessary, since
diagnosability is defined as a global property of the system.  In
order to deal with non-diagnosable systems, a more fine-grained, local
notion of \emph{trace diagnosability} is introduced, where
diagnosability is localized to individual traces.  This notion extends
the results on diagnosability checking from \cite{Cimatti2003}.

The second aspect is the \emph{maximality} of the diagnoser, that is,
the ability of the diagnoser to raise an alarm as soon as possible and
as long as possible, without violating the correctness condition.

The pattern-based language defined in~\cite{TACAS14,LMCS} is based on
temporal logic. In particular, the patterns are provided with an
underlying formal semantics expressed in epistemic temporal
logic~\cite{HalpernVardi1989}, where the \emph{knowledge} operator is
used to express the certainty of a condition, based on the available
observations.  The language is called Alarm Specification Language
with Epistemic operators ($ASL_ K$).  Diagnosis conditions and alarm
conditions are formalized using LTL with past operators, whereas the
definitions of trace diagnosability and maximality require epistemic
logic. The full specification, covering the concepts of (system and trace)
diagnosability and maximality, is shown in Figure~\ref{fig:kasl}.

\begin{figure*}[t]
\center
\scalebox{0.85}{%
\begin{tabular}{|l|l||l||l|} \hline
& Template &  $\maximality{Maximality} = False$& $\maximality{Maximality} = True$  \\
\hline \hline
\parbox[t]{5mm}{\multirow{3}{*}{\rotatebox[origin=c]{90}{$\diagnosability{Diag} = \dGlobal$ }}}
& \multirow{2}{*}{\textsc{ExactDel}}
& 
  $\correctness{G (\trigger{A} \rightarrow {Y^{n} \beta})} \wedge
  \completeness{G (\beta \rightarrow X^n \trigger{A})}$
& $\correctness{G (\trigger{A} \rightarrow {Y^{n} \beta})} \wedge
  \completeness{G (\beta \rightarrow X^n \trigger{A})} \ \wedge$ \\
& & 
& $\maximality{G (\strong{K Y^n \beta} \rightarrow \trigger{A})} $\\ \cline{2-4}
& \multirow{2}{*}{\textsc{BoundDel}}
& 
  $\correctness{G (\trigger{A} \rightarrow {O^{\le n} \beta})} \wedge
   \completeness{G (\beta \rightarrow F^{\le n} \trigger{A})} $
& $\correctness{G (\trigger{A} \rightarrow {O^{\le n} \beta})} \wedge
  \completeness{G (\beta \rightarrow F^{\le n} \trigger{A})} \ \wedge$ \\
& & 
& $ \maximality{G (\strong{K O^{\le n} \beta} \rightarrow \trigger{A})}$
  \\ \cline{2-4}
& \multirow{2}{*}{\textsc{FiniteDel}}
& 
  $\correctness{G (\trigger{A} \rightarrow {O \beta})} \wedge
   \completeness{G (\beta \rightarrow F \trigger{A})}$
& $\correctness{G (\trigger{A} \rightarrow {O \beta})} \wedge
   \completeness{G (\beta \rightarrow F \trigger{A})} \ \wedge$ \\
& & 
&$\maximality{G (\strong{K O \beta} \rightarrow \trigger{A})} $ \\ \cline{2-4}
\hline \hline
\parbox[t]{5mm}{\multirow{3}{*}{\rotatebox[origin=c]{90}{ $\diagnosability{Diag} =
      \dLocal$ \ \ \ \  }}}
& \multirow{2}{*}{\textsc{ExactDel}}
& $\correctness{G (\trigger{A} \rightarrow {Y^{n} \beta})} \ \wedge$
& $\correctness{G(\trigger{A} \rightarrow {Y^{n} \beta})} \ \wedge$ \\
& & $\completeness{G( \diagnosability{(\beta \rightarrow X^n \strong{K Y^n \beta} )}
  \rightarrow (\beta \rightarrow X^n \trigger{A}))}$
& $\completeness{G( \diagnosability{(\beta \rightarrow X^n \strong{K Y^n \beta} )}
  \rightarrow (\beta \rightarrow X^n \trigger{A}))} \ \wedge$ \\
& & & $\maximality{G (\strong{K Y^n \beta} \rightarrow \trigger{A})}$ \\
\cline{2-4}
& \multirow{2}{*}{\textsc{BoundDel}}
& $\correctness{G (\trigger{A} \rightarrow {O^{\le n} \beta})}\ \wedge$
& $\correctness{G (\trigger{A} \rightarrow {O^{\le n} \beta})}\ \wedge$ \\
& & $\completeness{G( \diagnosability{(\beta \rightarrow F^{\le n} \strong{K O^{\le n} \beta} )}
    \rightarrow (\beta \rightarrow F^{\le n} \trigger{A}))}$
& $\completeness{G( \diagnosability{(\beta \rightarrow F^{\le n} \strong{K O^{\le n} \beta} )}
 \rightarrow (\beta \rightarrow F^{\le n} \trigger{A}))} \ \wedge $ \\
& & & $\maximality{G(\strong{K O^{\le n} \beta} \rightarrow \trigger{A})}$ \\
\cline{2-4}
& \multirow{2}{*}{\textsc{FiniteDel}}
& $\correctness{G (\trigger{A} \rightarrow {O \beta})}\ \wedge $
& $\correctness{G (\trigger{A} \rightarrow {O \beta})}\ \wedge $ \\
& & $\completeness{G( \diagnosability{(\beta \rightarrow F \strong{K O \beta} )}
  \rightarrow (\beta \rightarrow F \trigger{A}))} $
& $\completeness{G( \diagnosability{(\beta \rightarrow F \strong{K O \beta} )}
  \rightarrow (\beta \rightarrow F \trigger{A}))} \ \wedge$ \\
& & & $\maximality{G(\strong{K O \beta} \rightarrow \trigger{A})} $ \\
\cline{2-4}
\hline
\end{tabular}}
\caption{$ASL_K$ specification patterns. Color key: cyan for diagnosability, red for maximality, orange for correctness, yellow for completeness.}
\label{fig:kasl}
\end{figure*}

The formalization encodes properties such as \emph{alarm correctness}
(whenever an alarm is raised by the FDI component, then the associated
condition did occur), and \emph{alarm completeness} (if an alarm is
not raised, then either the associated condition did not occur, or it
would have been impossible to detect it, given the available
observations). Alternative approaches that define diagnosability as
epistemic properties include \cite{Ezekiel2011} and \cite{Huang2013},
where the latter extends the definition of diagnosability to a
probabilistic setting. However, these works focus on finite-delay
diagnosability only, and do not consider the notion of trace
diagnosability.

The framework described in~\cite{TACAS14,LMCS} covers several
verification and validation problems.  The validation problem aims to
check whether the requirements capture the desired behaviors and
exclude unwanted ones.  Known techniques for requirements
validation~\cite{CRST12} include checking their consistency, their
compatibility with some possible scenarios, whether they entail some
expected properties and whether they are realizable. The verification
problem, on the other hand, checks whether a candidate diagnoser
fulfills a given set of requirements. These checks can be done using a
model checker for temporal epistemic logic such as MCK~\cite{mck} or,
if the specification falls in the pure LTL fragment, using a model
checker such as NuSMV~\cite{nusmv}.  The framework, finally, addresses the
problem of automated synthesis of a diagnoser from a given
specification. The idea is to generate an automaton that encodes the
set of possible states (called {\em belief states}) that represent the
estimation of the state of the system after each observation. Each
belief state of the automaton is annotated with the alarms that are
satisfied in all the states of the belief state. The algorithm
resembles the construction by Sampath~\cite{Sampath96} and
Schumann~\cite{Schumann2004}. It also extends the results of
\cite{Jiang2001}, which did not consider maximality and trace
diagnosability. Finally, we mention the problem of synthesizing
observability requirements, i.e. automatically discovering a set of
observations that is sufficient to guarantee diagnosability. This
problem is investigated in~\cite{aaai12}, which also addresses the
issue of synthesizing cost-optimal sets of observations.

The framework has been evaluated in the AUTOGEF~\cite{AUTOGEF-DASIA}
and FAME~\cite{FAME-DASIA,IMBSA-regular} projects, funded by the
European Space Agency, on a case study based on the EXOMARS Trace Gas
Orbiter.

\section{Timed Failure Propagation Graphs}
\label{sec:tfpg}
Classical safety assessment techniques such as FTA and FMEA do not
have a comprehensive support for analyzing the timing of failure
propagations, and make it difficult to obtain a global integrated
picture of the overall failure behavior of a system.  This in turn
makes it difficult to check whether a given FDIR architecture is able
to handle all possible faults and their propagation effects. To
address these issues, \emph{Timed Failure Propagation Graphs} (TFPGs)
~\cite{misra1992diagnosability,abdelwahed2009practical} have been
recently investigated as an alternative framework for failure
analysis.

TFPGs are labeled directed graphs that represent the propagation of
failures in a system, including information on timing delays and mode
constraints on propagation links. TFPGs can be seen as an abstract
representation of a corresponding dynamic system of greater
complexity, describing the occurrence of failures, their local
effects, and the corresponding consequences over time on other parts
of the system.  TFPGs are a very rich formalism: they allow to model
Boolean combinations of basic faults, intermediate events, and
transitions across them, possibly dependent on system operational
modes, and to express constraints over timing delays. In a nutshell,
TFPGs integrate in a single artifact several features that are
specific to either FMEA or FTA, enhanced with timing information.

\begin{figure}[t]
\centering
\includegraphics[width=0.50\textwidth]{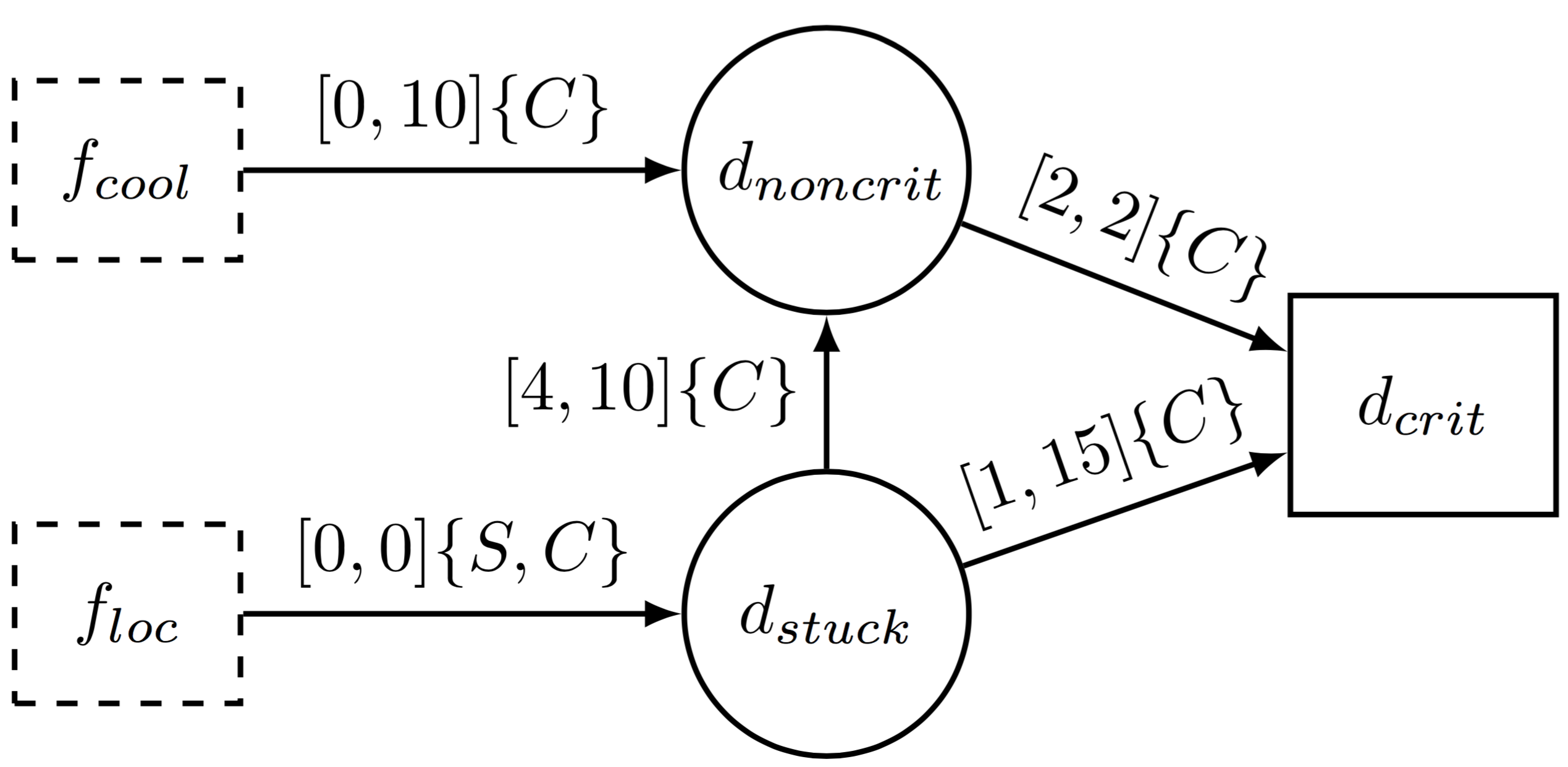}
\caption{An example TFPG. Dotted boxes are failure mode nodes, solid
  boxes discrepancy \andnode nodes, and circles discrepancy \ornode
  nodes.}
\label{fig:tfpg}
\end{figure}

An example TFPG is shown in Figure~\ref{fig:tfpg}.  Nodes represent
failure modes and discrepancies (i.e., failure effects). Edges model
the temporal dependency between the nodes; they are labeled with lower
and upper bounds on the propagation delay, and with labels indicating
the system modes where the corresponding propagations are possible.
The semantics of
TFPGs~\cite{bittner2016automated,bittner2016synthesis} is such that a
node is activated when a failure propagation has reached it. An edge
is active whenever the source node is active and the system mode
enables the propagation. Discrepancies include both \ornode and
\andnode nodes. In the former case, any of the incoming edges may
trigger the propagation, whereas in the latter case all of them must
be active. If an edge is deactivated during the propagation, due to
mode switching, the propagation stops.

TFPGs have been investigated in the frame of the FAME
project~\cite{FAME-DASIA,IMBSA-regular}, funded by the European Space
Agency (ESA). Here, a novel, model-based, integrated process for FDIR
design was proposed, which aims at enabling a consistent and timely
FDIR conception, development, verification and validation. More
recently,
~\cite{bozzano2015smt,bittner2016automated,bittner2016synthesis} have
investigated TFPG-based validation and formal analyses. In particular,
~\cite{bozzano2015smt} focuses on the validation of TFPGs, seen as
stand-alone models, using Satisfiability Modulo Theories (SMT)
techniques. Validation includes several criteria, such as possibility,
necessity and consistency checks, and TFPG refinement.
\cite{bittner2016automated} addresses TFPG validation (called {\em
  behavioral validation}), and tightening of TFPG delay bounds, with
respect to a system model of reference. In this context, the TFPG is
seen as an abstract version of the system model, and it is possible to
check whether the TFPG is complete, i.e. it represents all behaviors
that are possible in the system. Behavioral validation is performed by
discharging a set of proof obligations, using either a model checker
for metric temporal logic, or by reduction to LTL model checking.
Finally, ~\cite{bittner2016synthesis} develops algorithms for the
automatic synthesis of a TFPG from a reference system model. The
generated TFPG is guaranteed to be complete with respect to the system
model. Graph synthesis is carried out by using model checking routines
to compute sets of MCS and by simplifying the resulting graph by means
of a set of static rules. Parameter synthesis techniques are used for
edge tightening.

TFPGs have also been applied to several case studies in the context of
an ESA deep-space probe design~\cite{bittner2016thesis}, whose mission
is characterized by high requirements on on-board autonomy and
FDIR. In particular, TFPGs have been applied to the failure analysis
of the ``Solar Orbiter'' (SOLO) satellite.

\section{Conclusions}
\label{sec:conclusions}
In this paper, we have reviewed some recent work on the
design of fault management systems. The concepts of causality and
temporal dependencies that arise in this setting have similarities
with classical theories of causality, such as counterfactual
causality~\cite{halpern2005causes,Halpern2}. Such theories are defined
using structural equations, but can be readily re-formulated for
transition systems~\cite{Leue1,Leue2}. A thorough investigation of the
implications of causality theories in the context of fault management
systems is part of our future work. We outline here some related work
and possible directions for future investigation.

The notion of causality in FTA closely resembles the idea of
identifying minimal sets of (necessary and sufficient) causes as in
classical causality theories. However, given an effect (TLE), FTA is
interested in such sets of causes (i.e., faults -- identified
beforehand) in all possible scenarios, whereas classical theories
focus on identifying the causes in a given scenario of
interest. Moreover, in FTA a cause (i.e., MCS) need not be a
sufficient condition -- sometimes an additional condition on the
environment might be needed. Such condition resembles the notion on
{\em contingency} in causality theories, and could be represented
using FTA gates (e.g., a pair inhibit gate/conditioning
event~\cite{FTH2}). In~\cite{Leue1,Leue2}, causality is extended to
encompass the notions of ordering and non-occurrence of events. This
approach extends the ordering analysis proposed
in~\cite{bozzano:02,DBLP:journals/cj/BozzanoCKNNR11}.

FDI logic links effects with causes, similar to classical causality
theories, but using {\em observables} only. An alarm, in this context,
corresponds to an effect or, more precisely, to a signal which is
triggered by the detection/identification of a given effect. Given a
fault $F$ and an alarm $A$, FDI correctness implies that $F$ is (part
of) a cause of $A$, whereas FDI completeness does not necessarily
imply that $F$ is a cause of $A$, since false alarms are
possible. However, correctness and completeness together imply that
$F$ is the (unique) cause of $A$. Finally, diagnosability is related
to the realizability of FDI logic, and trace diagnosability
corresponds to diagnosability in a specific scenario.

Finally, TFPG analyses have similaries with FTA -- in fact, TFPG
synthesis is built on top of MCS computation. However, TFPGs are more
expressive multi-node networks, enriched with time bounds and modes,
and nodes may include dependent effects (dicrepancies). A propagation
in a TFPG is necessary, in the sense that a discrepancy activation
implies the propagation through at least one input propagation link,
whereas a propagation is inevitable in the sense that a propagation
implies the activation of the correspondent discrepancy. Inevitability
may be enforced using time bounds and/or modes.

As part of our future work, we want to analyze more closely the
difference between causality and temporal dependencies/temporal
correlation.  In some scenarios of interest, motivated by practical
case studies, it appears that temporal correlation between causes and
different effects of the same cause, may lead to identifying a
temporal-correlated effect as part of the causes of the effect of
interest. Distinguishing causality from temporal correlation would
require going beyond the trace-based semantics. We are currently
looking for meaningful and sound definitions that can encompass such
cases.

\paragraph*{Acknowledgments}
The results presented in this paper are a joint work with several
people, including Benjamin Bittner, Alessandro Cimatti, Marco Gario
and Stefano Tonetta.

\bibliographystyle{eptcs}
\bibliography{refs}

\end{document}